\documentclass[runningheads]{llncs}

\usepackage{graphicx}
\usepackage{amsmath,amssymb} 
\usepackage{color}
\usepackage[width=122mm,left=12mm,paperwidth=146mm,height=193mm,top=12mm,paperheight=217mm]{geometry}

\usepackage[lofdepth,lotdepth,caption=false]{subfig}

\usepackage{booktabs}
\usepackage{array}

\DeclareGraphicsExtensions{.pdf,.jpg,.png}

\begin{document}
\pagestyle{headings}
\mainmatter

\title{Amodal Instance Segmentation} 

\titlerunning{Amodal Instance Segmentation}

\authorrunning{Ke Li and Jitendra Malik}

\author{Ke Li and Jitendra Malik}


\institute{Department of Electrical Engineering and Computer Sciences,\\
	University of California, Berkeley\\
	\email{ \{ke.li,malik\}@eecs.berkeley.edu}
}

\maketitle

\begin{abstract}
We consider the problem of amodal instance segmentation, the objective of which is to predict the region encompassing both visible and occluded parts of each object. Thus far, the lack of publicly available amodal segmentation annotations has stymied the development of amodal segmentation methods. In this paper, we sidestep this issue by relying solely on standard modal instance segmentation annotations to train our model. The result is a new method for amodal instance segmentation, which represents the first such method to the best of our knowledge. We demonstrate the proposed method's effectiveness both qualitatively and quantitatively. 
\end{abstract}

\section{Introduction}

\begin{figure}
    \centering
    \includegraphics[width=1.0\textwidth]{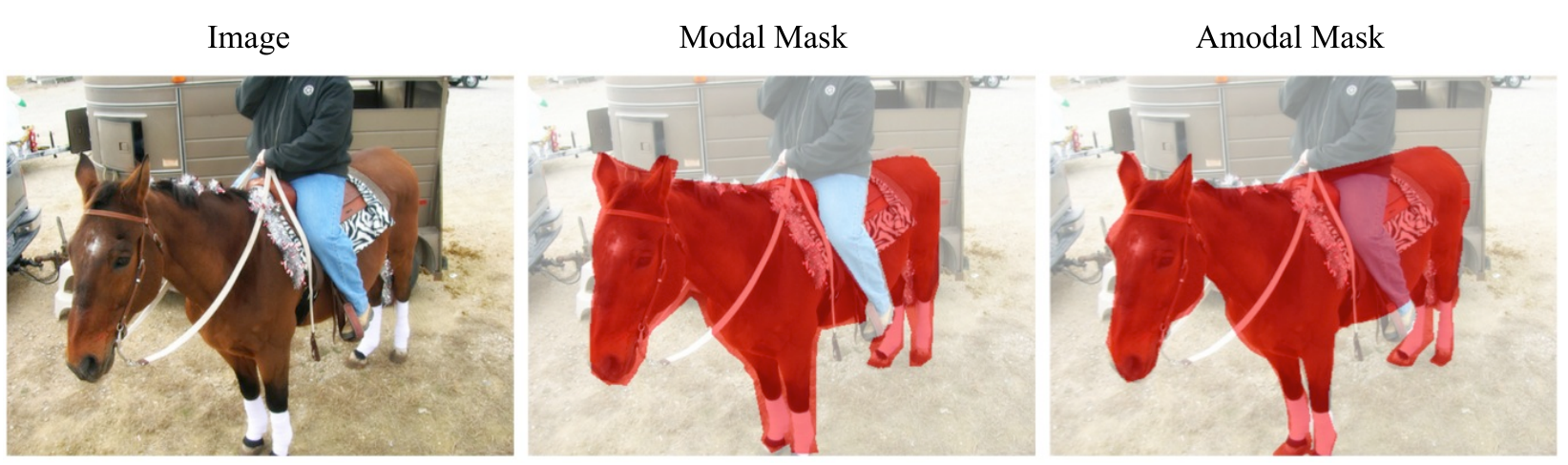}
    \caption{Target outputs for modal and amodal segmentation}
    \label{fig:task}
\end{figure}

Consider the horse shown in the left panel of Fig. \ref{fig:task}. The task of instance segmentation requires marking the visible region of the horse, as shown in the middle panel, and has been tackled by several existing algorithms~\cite{sds,dai2015convolutional,hypercolumn,iis,dai2015instance}. In this paper, we consider a different task, which requires marking both the visible and the occluded regions of the horse, as shown in the right panel. In keeping with terminology used in the psychology literature on visual perception~\cite{kanizsa1979organization}, we refer to the former task as \emph{modal instance segmentation} and the latter task as \emph{amodal instance segmentation}. 

A natural question to ask is if the task of amodal instance segmentation is well-posed: given only the visible portions of an object, there are many possible configurations of the hidden portions of the object, all of which appear to be plausible hypotheses to a human. This is particularly true for articulated objects under heavy occlusion. For example, if the lower body of a person is blocked from view, there is no single correct hypothesis for the configuration of the person's legs -- the person could be sitting or standing, and so hypotheses consistent with either pose would be equally valid. Despite this ambiguity, humans are capable of performing amodal completion and tend to predict the occluded regions with high degrees of consistency~\cite{sas}. 

An amodal segmentation system would open the way to sophisticated occlusion reasoning. For instance, given an amodal segmentation mask, we can infer the presence, extent, boundary and region of occlusions by comparing it to the modal segmentation mask. We can also deduce the relative depth ordering by comparing the modal and amodal masks of the occluded and occluding objects. The information derived from the amodal segmentation mask can be further used downstream for a variety of interesting applications. For instance, we can estimate the physical dimensions of an object in the real world using its amodal bounding box, as demonstrated by Kar et al.~\cite{kar2015amodal}

The fact that all these occlusion reasoning problems can be reduced to amodal instance segmentation implies that amodal instance segmentation is more challenging than all these problems combined. An amodal segmentation system must not only be capable of determining if an object is occluded, but also where it is occluded. It must be able to hypothesize the shape of the occluded portion even though it has never seen the whole object before. It must be sensitive enough to detect the diminished signal from the small part of the occluded object that remains visible, but must be robust enough to avoid being misled by strong signals from occluding objects. 

Furthermore, additional complicating factors arise if one were to attempt training a model for the task, the chief among them being the lack of supervised training data. While efforts are underway to collect amodal segmentation annotations~\cite{sas}, no amodal segmentation data is publicly available at the time of writing. In the meantime, we need to devise a clever way of using existing data to train a model for this new task. 

In this paper, we present a new method for amodal instance segmentation, which to the best of our knowledge is the first such method. We train our method purely from existing modal instance segmentation data, thereby sidestepping issues arising from the lack of supervised training data. We make a key observation: while it is not possible to compute the amodal mask of an object from the modal mask by undoing occlusion, it is easy to do the reverse. Instead of undoing existing occlusion, we add synthetic occlusion and retain the original mask, which essentially becomes the true amodal mask for the composite image. We train a convolutional neural net to recover the original mask from the generated composite image. We do not assume knowledge of the amodal bounding box at test time; instead, we infer it from the amodal segmentation heatmap using a new strategy, which we dub Iterative Bounding Box Expansion. We demonstrate that despite being trained on synthetic data, the resulting model is quite effective at predicting amodal masks on images with real occlusions. 

\section{Related Work}

Efforts toward understanding the semantic meaning associated with free-form regions in images started with work on figure-ground segmentation, the objective of which is to identify the foreground pixels in typically object-centric images. Early methods~\cite{borenstein2002class,yu2003object,leibe2004combined,kumar2005obj,levin2006learning} investigated ways of combining top-down and bottom-up segmentation approaches and incorporating class-specific or object-specific templates of the foreground object's appearance. Later on, the focus shifted to the more general problem of semantic segmentation, which aims to identify the pixels that belong to each object category in more complex images. A diverse range of approaches for this problem have been developed; earlier approaches extend CRF-based formulations~\cite{ladicky2010graph,kohli2010energy,boix2012harmony}, consider combining object detections with region proposals~\cite{gu2009recognition}, scoring groupings of over-segmented regions~\cite{vijayanarasimhan2011efficient}, aggregating information from multiple foreground-background hypotheses~\cite{carreira2012object} and synthesizing scores from different overlapping regions to obtain a pixel-wise classification~\cite{arbelaez2012semantic}. Later approaches~\cite{farabet2013learning,pinheiro2014recurrent,long2015fully,zheng2015conditional} use feedforward or recurrent neural net models to extract features or to predict the final label of each pixel directly from images. In effort at achieving understanding of a scene at a finer level of granularity, recent work has focused on the task of instance segmentation, the goal of which is to identify the pixels that belong to each individual object instance. The predominant framework for this task is to find the bounding box of each instance using an object detector and predict the figure-ground segmentation mask inside each box. Earlier approaches rely on DPM~\cite{felzenszwalb2010object} detections and predict segmentation masks using a simple appearance model~\cite{yang2012layered,parkhi2011truth,dai2012learning}, combine DPM detections and semantic segmentation predictions~\cite{fidler2013bottom,dong2014towards} or adopt a transductive approach~\cite{tighe2014scene}. More recent methods leverage the power of convolutional neural nets. SDS~\cite{sds} and Dai et al.~\cite{dai2015convolutional} use a neural net to compute features on region proposals and classifies them using an SVM, while the Hypercolumn net~\cite{hypercolumn}, Iterative Instance Segmentation~\cite{iis} and Multi-Task Network Casades~\cite{dai2015instance} predict the segmentation mask directly from the image patch. We view the task of amodal instance segmentation as the natural next step in this direction. 

There has been relatively little work exploring amodal completion. Kar et al.~\cite{kar2015amodal} tackled the problem of predicting the amodal bounding box of an object. Gupta et al.~\cite{gupta2013perceptual} explored completing the occluded portions of planar surfaces given depth information. To the best of our knowledge, there has been no algorithmic work on general-purpose amodal segmentation. However, there has been work on collecting amodal segmentation annotations. Zhu et al.~\cite{sas} collected amodal segmentation annotations on BSDS images, but has yet to make them publicly available. As far as we know, the proposed method represents the first method for amodal segmentation. 

\section{Generating Training Data}

\begin{figure}
    \centering
    \includegraphics[width=1.0\textwidth]{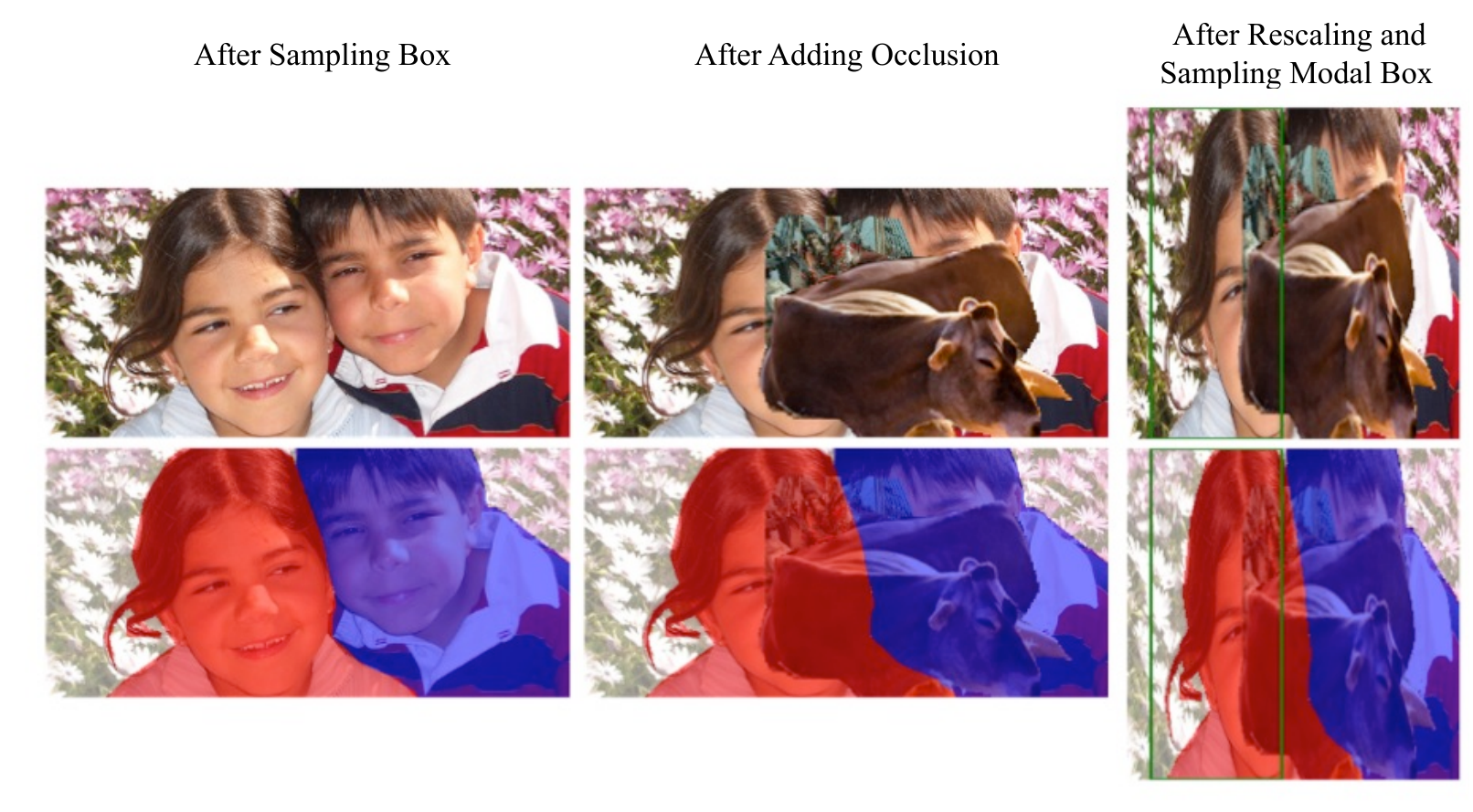}
    \caption{The image patch and the target segmentation mask after each step of the sampling procedure. Red regions in the segmentation mask are assigned the positive label, white regions in the segmentation mask are assigned the negative label and blue regions in the segmentation mask are assigned the unknown label. The green box denotes the jittered modal bounding box}
    \label{fig:sampling_process}
\end{figure}

\subsection{Overview}

We generate amodal training data solely from standard modal instance segmentation annotations. In our case, we use the Semantic Boundaries (SBD) annotations~\cite{sbd} on the PASCAL VOC 2012 \textit{train} set as the data source. We generate three types of data: image patches, modal bounding boxes and target segmentation masks. Image patches and modal bounding boxes are used as input to the model and target segmentation masks are used as supervisory signal on the output of the model. 

The key observation we leverage is that the phenomenon of occlusion can be easily simulated by overlaying objects on top of other objects. More concretely, we first generate randomly cropped image patches that overlap with at least one foreground object instance, which we will refer to as the main object. We then extract random object instances from other images and overlay them on top of the randomly cropped patches with their modal segmentation masks serving as the alpha matte. Each overlaid object is positioned and scaled randomly in a way that ensures a moderate degree of overlap with the main object. Essentially, this procedure generates composite patches where the main object is partially occluded by other objects. 

Next, for each composite patch, we find the smallest bounding box that encloses the portions of the main object that remain visible. This is essentially ground truth modal bounding box of the main object in the composite patch. To simulate noisy modal localization at test time, we jitter the bounding box randomly. 

Finally, we generate the target segmentation masks corresponding to the composite patches produced above. For each patch, we take the corresponding part of its original modal segmentation mask and label the pixels belonging to the object as positive, pixels belonging to the background as negative and pixels belonging to other objects as unknown. This manner of label assignment captures what we know about the amodal mask given the modal mask -- we know the visible portion of the object must be a part of the whole object and that the object cannot be occluded by the background. However, the object \emph{may} be occluded by other objects in the image; consequently, the pixels belonging to other objects in the modal mask are labelled as unknown. Because the original modal mask is not affected by overlaid objects, this mask includes portions of the main object that were originally visible but are now occluded in the composite patch. Hence, the target mask is consistent with the true amodal mask. 

\subsection{Implementation Details}

\begin{figure}
    \centering
    \includegraphics[width=1.0\textwidth]{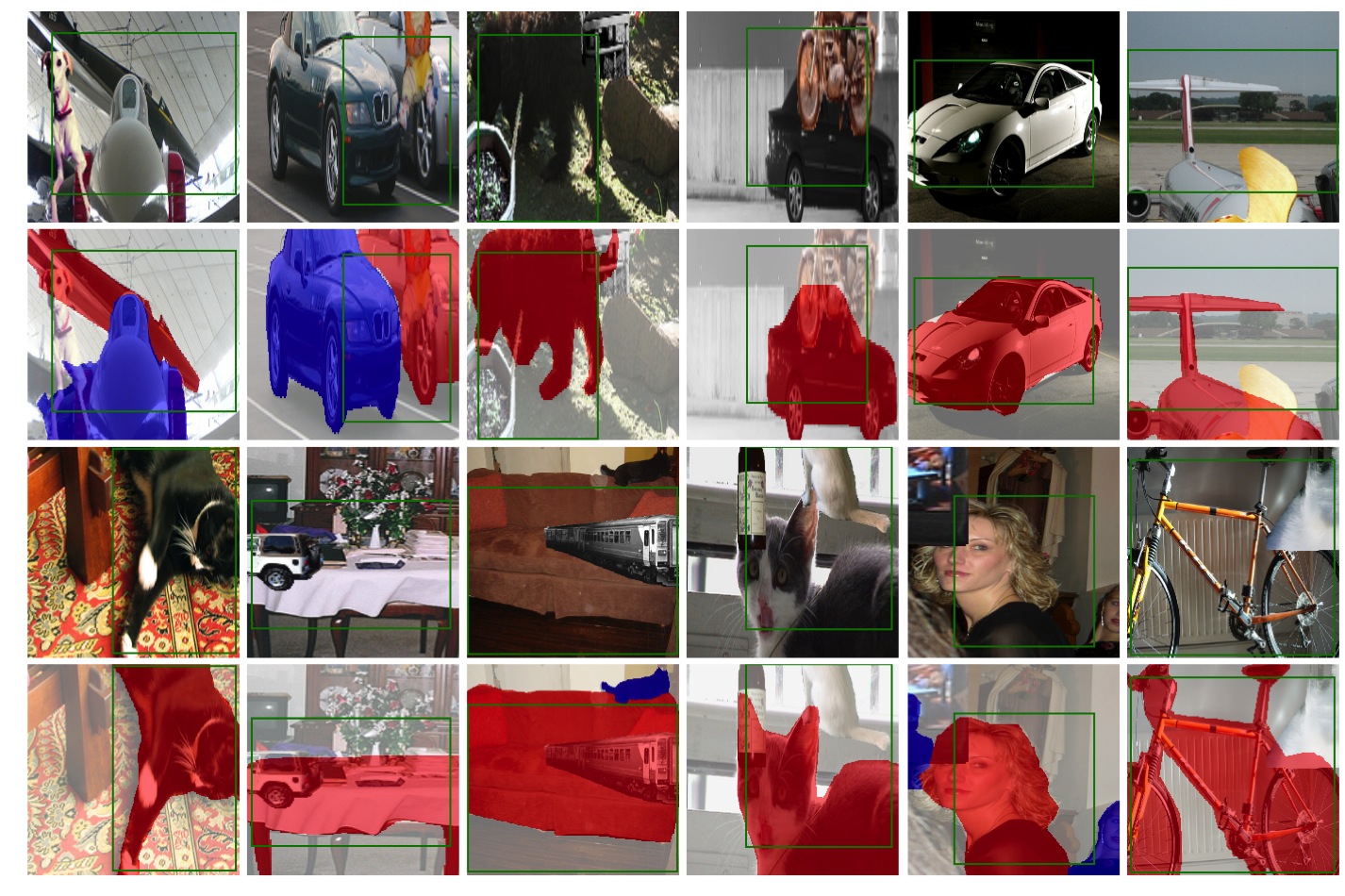}
    \caption{Random samples from the generated training data}
    \label{fig:train_samples}
\end{figure}

Data is generated on-the-fly during training. To generate a training example, we sample an image uniformly and then sample an object instance from the image, which will be referred to as the main object. We then randomly sample a bounding box that overlaps with the main object's bounding box along each dimension by at least 70\%. The size of the sampled box is randomly chosen and the length of each dimension is between 70\% and 200\% of the length of the corresponding dimension of the object's bounding box. Next, we choose the number of objects to overlay onto the patch inside the bounding box randomly by picking an integer from 0 to 2. To select an object to overlay, we sample an image and then sample a random object instance from the image. The object is placed at a random location that overlaps with the main object and scaled randomly so that the length of its shortest dimension is 75\% of the length of the corresponding dimension of the patch on average. After each of the above operations, we check if the proportion of the main object that remains visible falls below 30\%. If it does, we undo the most recent operation and try again. Otherwise, we proceed to find the bounding box the encloses the portion of the main object that remains visible and randomly samples a box that overlaps with the bounding box by at least 75\% along each dimension and can differ in size from the bounding box by at most 10\% in each dimension. 

\section{Predicting Amodal Mask and Bounding Box}

\subsection{Testing}

We take the modal bounding box, that is, the bounding box of the visible part of the object, and the category of the object as given, which can be obtained from an object detector, like R-CNN~\cite{rcnn}, fast R-CNN~\cite{fast_rcnn} or faster R-CNN~\cite{faster_rcnn}. We then compute the modal segmentation heatmap using Iterative Instance Segmentation (IIS)~\cite{iis}, which is the state-of-the-art method for modal instance segmentation. 

The proposed algorithm proceeds to predict the amodal segmentation mask and bounding box in an iterative fashion using a new strategy that will be referred to as Iterative Bounding Box Expansion. Initially, the amodal bounding box is set to be the same as the modal bounding box. In each iteration, given the amodal bounding box, we feed the patch inside the amodal bounding box to a convolutional neural net, which predicts the amodal segmentation mask inside an expanded amodal bounding box that also includes areas immediately outside the original amodal bounding box. We compute the average heat intensity in the areas lie above, below, to the left and to the right of the original bounding box. If the average heat intensity associated with a particular direction is above a threshold, which is set to 0.1 in our experiments, we expand the bounding box in that direction and take this new bounding box to be the amodal bounding box used in the next iteration. This procedure is repeated until the average heat intensities in all directions are below the threshold. To obtain the final amodal segmentation mask, we colour in all pixels whose intensities in the corresponding heatmap exceed $0.7$. Similarly, we obtain the modal segmentation mask by thresholding the modal heatmap at $0.8$. 

\begin{figure}
    \centering
    \includegraphics[width=1.0\textwidth]{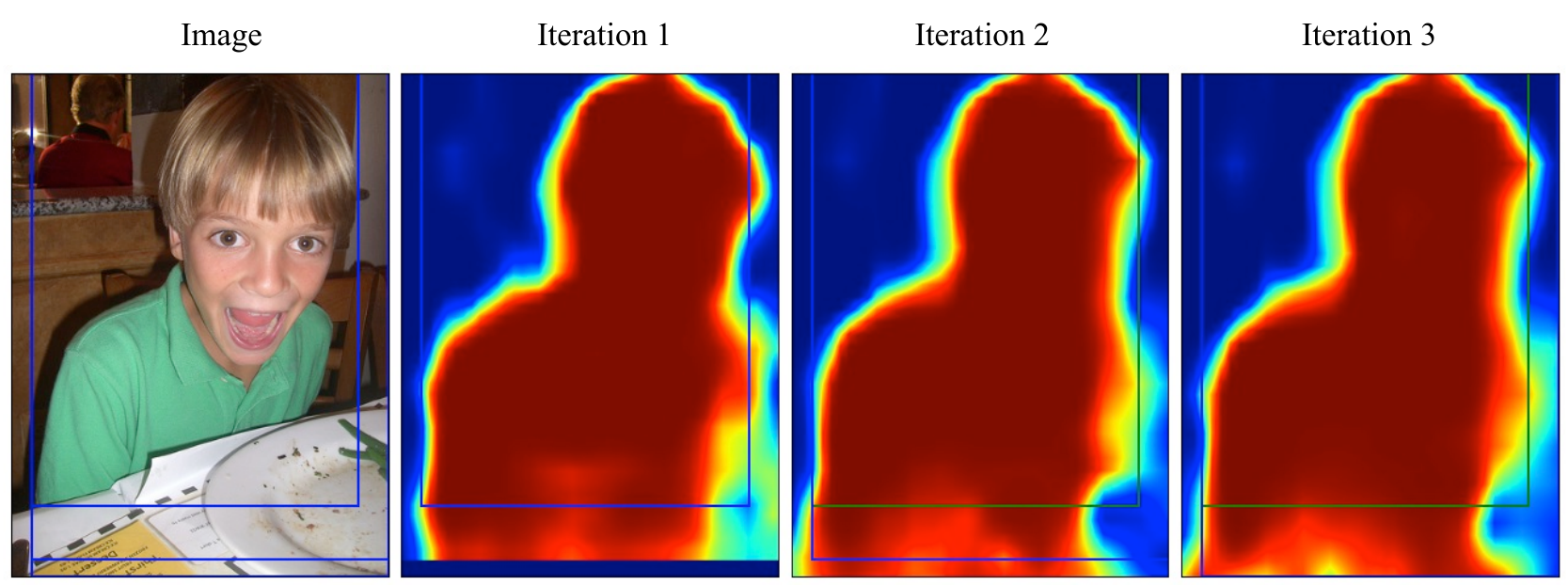}
    \caption{Amodal segmentation heatmap after each iteration of Iterative Bounding Box Expansion. The green box denotes the modal bounding box and the blue box denotes the amodal bounding box. In each iteration, we use the average heat intensity \emph{outside} the amodal bounding box to decide whether to expand the amodal bounding box in the next iteration}
    \label{fig:ibbe}
\end{figure}

\subsection{Training}

The convolutional neural net we use takes in an image patch, a modal segmentation heatmap and a category and outputs the amodal segmentation heatmap. The net has the same architecture as that used by IIS, which is based on the hypercolumn architecture introduced by Hariharan et al.~\cite{hypercolumn}. This architecture is designed to take advantage of both low-level image features at finer scales and high-level image features at coarser scales. It does so by making the final heatmap prediction dependent on the summation of upsampled feature maps from multiple intermediate layers, which is known as the hypercolumn representation. The version of the architecture we use is based on the VGG 16-layer net~\cite{vgg}, which is referred to as ``O-Net'' in \cite{hypercolumn}. The IIS architecture is a variant of this architecture that also takes in an initial heatmap hypothesis via an additional category-dependent channel as input, which can be set to constant if no initial heatmap hypothesis is available. If an initial heatmap hypothesis is provided, the model refines the heatmap hypothesis to produce an improved heatmap prediction. Using this architecture, IIS is able to iteratively refine its own heatmap prediction by feeding in its heatmap prediction from the preceding iteration as input. 

Each training example consists of an image patch, a modal bounding box and a target amodal segmentation mask. To prepare input to the net, we take the part of image patch that lies inside the modal bounding box and scale it anisotropically to $224 \times 224$, feed it to IIS as input. We take the modal segmentation heatmap produced by one iteration of IIS, align it to the coordinates of the original image patch and upsample it to $224 \times 224$ using bilinear interpolation. Because the model should predict the mask corresponding to an area larger than the image patch it sees, we remove 10\% of the image patch from each of the four sides and rescale it to $224 \times 224$. If less than 10\% of the pixels in this new patch belong to the visible portion of the object, we reject the current sample and generate a new training example. We centre the data by subtracting the mean pixel from the image patch and transforming the modal segmentation heatmap element-wise to lie between $-127$ and $128$. Finally, we feed the image patch, the modal segmentation heatmap and the target amodal mask to the model for training. 

The model is trained end-to-end using stochastic gradient descent with momentum on mini-batches of 32 patches starting from weights of the model used by IIS. The loss function we use is the sum of pixel-wise negative log likelihoods over all pixels with known ground truth labels. We apply instance-specific weights that are inversely proportional to the factor by which each patch is upsampled. We train the model with a constant learning rate of $10^{-5}$, weight decay of $10^{-3}$ and momentum of $0.9$ for $50,000$ iterations. 

\section{Experiments}

Because there is no existing dataset with amodal instance segmentation annotations, there is no ground truth against which the predictions can be evaluated, making it difficult to perform a quantitative evaluation. We first present qualitative results and then perform an indirect quantitative evaluation against coarse-level annotations on the full PASCAL VOC 2012 \textit{val} set. To perform a direct quantitive evaluation, we annotated $100$ randomly chosen occluded objects from the same dataset with amodal segmentation masks and evaluate the proposed method on this subset. 

\subsection{Qualitative Results}

\begin{figure}
    \centering
    \includegraphics[width=1.0\textwidth]{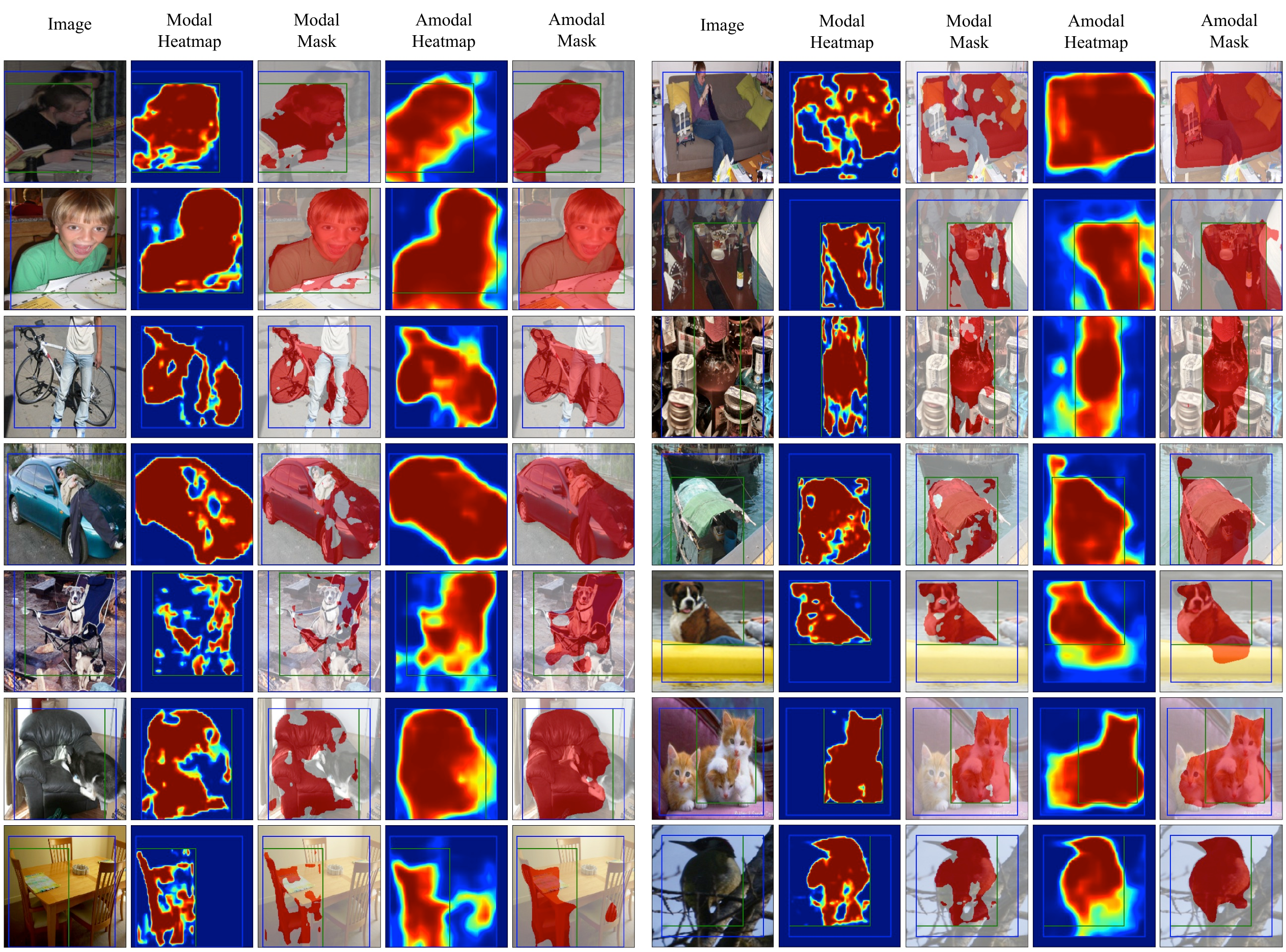}
    \caption{Examples of amodal segmentation mask predictions for occluded objects which we judge to be correct. The first column shows the part of the image containing the amodal and modal bounding box, with the green box denoting the original modal bounding box that is given and the blue box denoting the expanded amodal bounding box found by Iterative Bounding Box Expansion. If a side of the amodal bounding box is adjacent to the border of the image patch, the patch abuts the corresponding border of the whole image. The next four columns show the modal segmentation heatmap and mask produced by IIS and the amodal segmentation heatmap and mask produced by the proposed method}
    \label{fig:qual_results_successes}
\end{figure}

\begin{figure}
    \centering
    \includegraphics[width=1.0\textwidth]{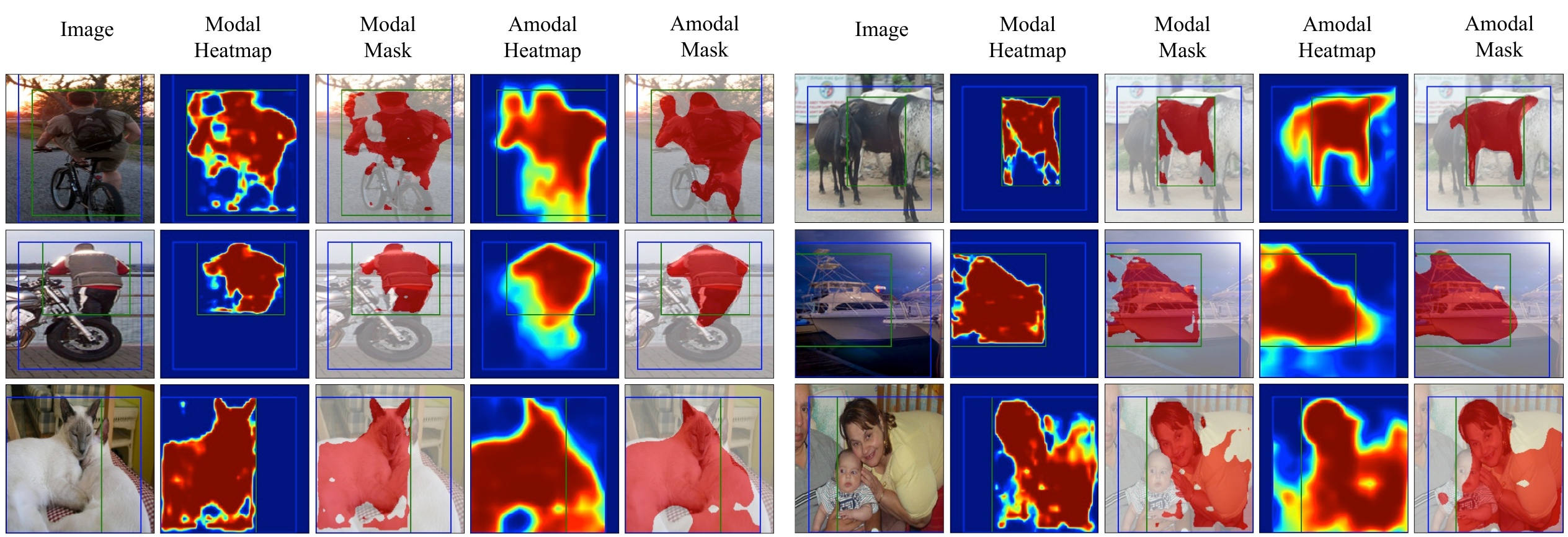}
    \caption{Examples of amodal segmentation mask predictions for occluded objects which we judge to be incorrect. The visualizations follow the same format as Fig. \ref{fig:qual_results_successes}}
    \label{fig:qual_results_failures}
\end{figure}

\begin{figure}
    \centering
    \includegraphics[width=1.0\textwidth]{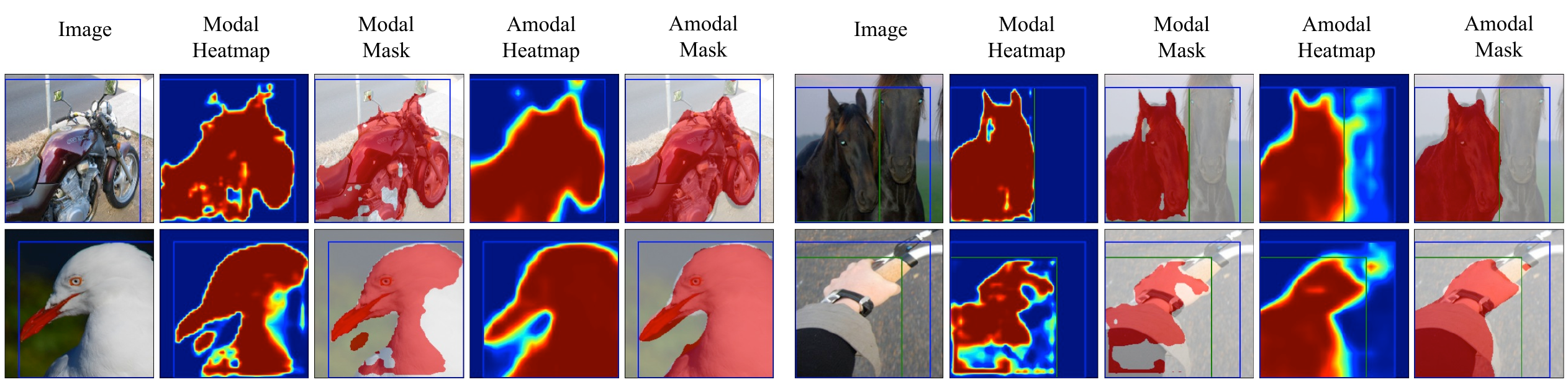}
    \caption{Examples of amodal segmentation mask predictions for unoccluded objects. The visualizations follow the same format as Fig. \ref{fig:qual_results_successes}}
    \label{fig:qual_results_unoccl}
\end{figure}

We use the proposed method to generate amodal segmentation mask predictions for objects in the PASCAL VOC 2012 \textit{val} set. Because the focus of this paper is on the segmentation system, we take the modal bounding box and the category of the object of interest as given and obtain them from the ground truth. We show the amodal heatmap and mask predictions produced by the proposed method Figures \ref{fig:qual_results_successes}, \ref{fig:qual_results_failures} and \ref{fig:qual_results_unoccl}, along with the modal heatmap and mask predictions generated by IIS for comparison. As shown, the proposed method is generally quite effective. In particular, even though the synthetic occlusions generated for training purposes do not appear entirely realistic, the proposed method is able to devise plausible hypotheses for hidden portions of objects caused by real occlusions. 

We classify occlusions into two types: cases where the occluding object is mostly contained inside the occluded object and cases where a significant portion of the occluding object lies outside the occluded object. We will refer to the former as interior occlusions and the latter as exterior occlusions. For interior occlusions, the goal is to predict the mask between visible portions of the object, whereas for exterior occlusions, the goal is to predict the mask beyond the visible portion. There is typically a single correct way to handle interior occlusions: if it results from a true occlusion, the corresponding hole should be filled in; otherwise, it should not be. Therefore, for these cases, the task is less ambiguous and relatively easy. On the other hand, there are generally multiple equally valid ways to handle exterior occlusions: there are many possible ways to extend the visible portion that all lead to plausible amodal masks. So, for these cases, the task is more ambiguous and challenging. The method need to decide how much to extend the modal mask by in every direction. To do so, it must rely on the knowledge it learns about the general shape of objects of the particular category to produce a plausible amodal mask hypothesis. 

We first examine examples of occluded objects for which the proposed method produces correct amodal segmentation masks, which are shown in Fig. \ref{fig:qual_results_successes}. As shown, the proposed method is able to successfully predict the amodal mask on images with interior or exterior occlusions. Surprisingly, on some images where the modal prediction is poor, like the image with a dog on a folding chair, the proposed method is able to produce a fairly good amodal mask. Finally, the proposed method is able to produce remarkably good amodal masks on some challenging images with exterior occlusions, such as the image depicting a dog on a kayak and the images in the bottom two rows, suggesting that the proposed method is able to learn something about the general shape of objects. 


Next, we take a look at examples of occluded objects for which the predicted amodal segmentation masks are incorrect, which are shown in Fig. \ref{fig:qual_results_failures}. The mistakes may be caused by the rarity of unusual poses in the training set, large variation in the plausible configurations of the occluded portions of the object, similarity in appearance of adjacent objects or erroneous modal predictions. 

While the preceding examples show that the model is capable of performing amodal completion, we need to make sure that the model does not hallucinate when presented with unoccluded objects. We explore some examples of unoccluded objects and the corresponding mask predictions in Fig. \ref{fig:qual_results_unoccl}. Since the objects are unoccluded, amodal masks should be the same as modal masks. As shown, the amodal predictions are similar to or more accurate than the modal predictions. This may be explained by the robustness acquired by an amodal segmentation model: by learning to be robust to occlusion, the model also learns to be robust to variations in low-level patterns in the image that may confuse a modal segmentation model. 

We include additional examples of heatmap and mask predictions in the supplementary material. 

\subsection{Indirect Evaluation}

We use the proposed method combined with a modal instance segmentation method like IIS to predict the presence or absence of occlusion. We do so by predicting the modal and amodal segmentation masks for each object in the PASCAL VOC 2012 \textit{val} set and computing the following ratio:
\[
\frac{\mbox{area}\left(\textit{modal mask}\cap\textit{amodal mask}\right)}{\mbox{area}\left(\textit{amodal mask}\right)}
\]
Intuitively, this ratio measures the degree by which an object is occluded. For an unoccluded object, because the amodal mask should be the same as the modal mask, this ratio should be close to $1$. On the other hand, for a heavily occluded object, only a small proportion of the pixels inside the amodal mask should also be included in the modal mask; as a result, this ratio should be significantly less than $1$. We will henceforth refer to this ratio as the \emph{area ratio}. 

\begin{figure}
    \centering
    \subfloat[]{
        \includegraphics[width=0.43\textwidth]{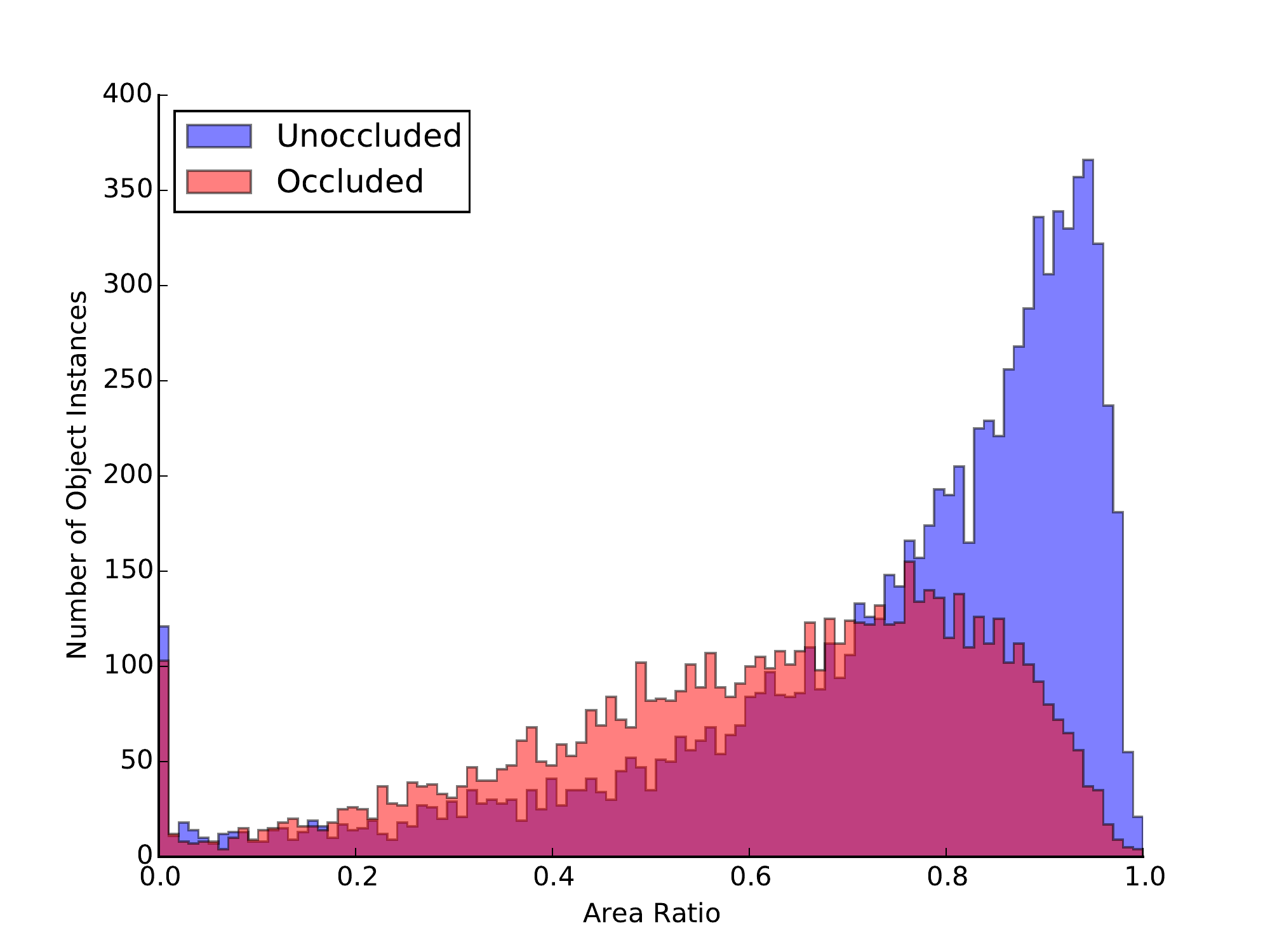}
        \label{fig:area_ratio_histogram}
    }
    \subfloat[]{
        \includegraphics[width=0.43\textwidth]{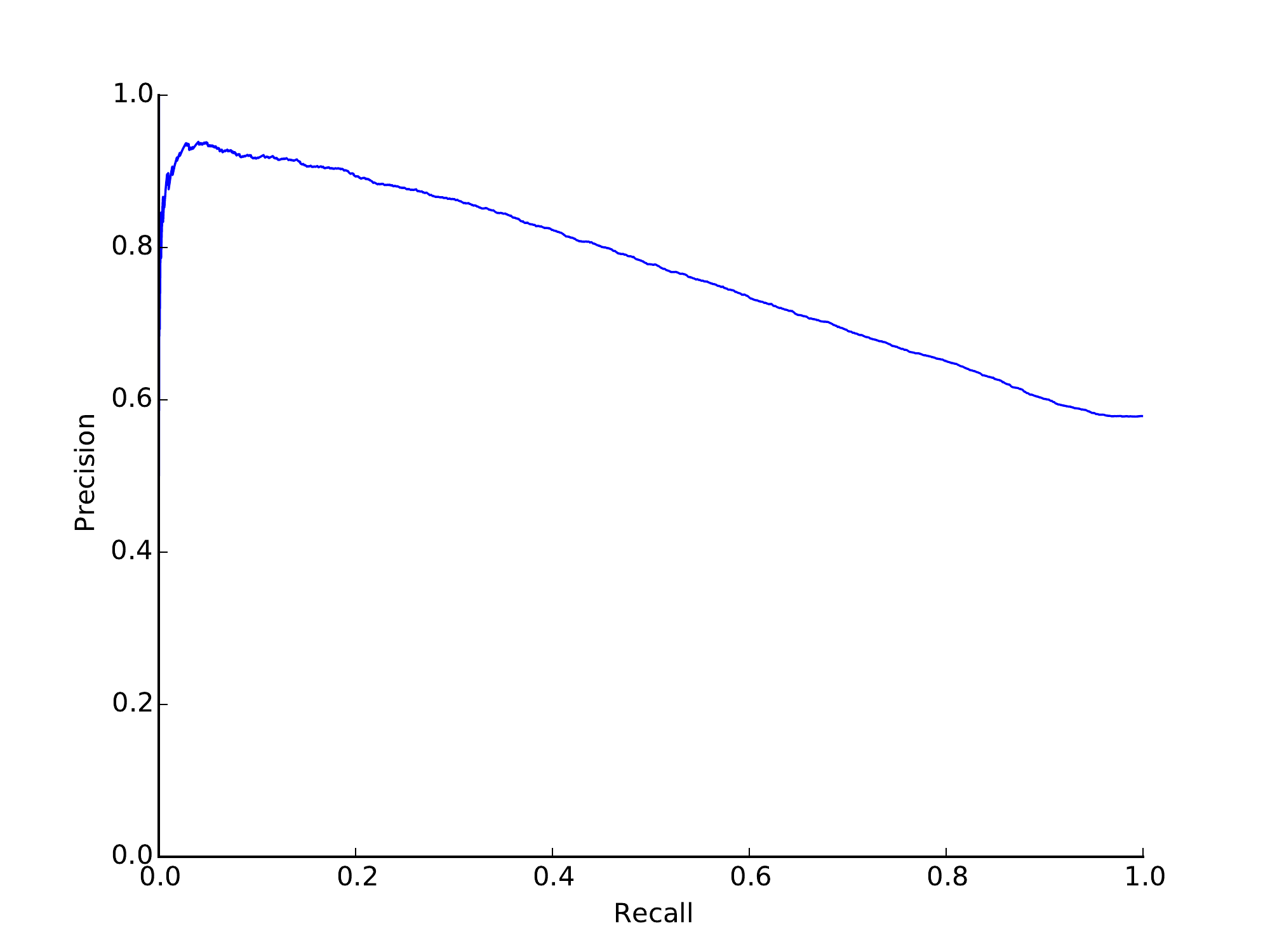}
        \label{fig:area_ratio_pr_curve}
    }
    \caption{(a) Distribution of area ratios of occluded and unoccluded objects in the PASCAL VOC 2012 \textit{val} set. The range of possible area ratios is discretized into 100 bins of equal width and the vertical axis shows the number object instances whose area ratios lie in a particular bin. (b) The precision-recall curve for predicting the absence of occlusion by thresholding the area ratio}
\end{figure}

We compare our predictions to the occlusion presence annotations in the PASCAL VOC 2012 \textit{val} set, which are available for all object instances and specify whether they are occluded. First, we compute the modal and amodal mask predictions using IIS and the proposed method for all objects that are annotated as being occluded and plot the distribution of area ratios. We then do the same for objects that are annotated as being unoccluded. These two distributions are shown in Fig. \ref{fig:area_ratio_histogram}. As shown, the distribution for unoccluded objects is heavily skewed towards high area ratios, whereas the distribution for occluded objects peaks at an area ratio of around $0.75$. This indicates the predicted amodal masks for occluded objects typically have more pixels outside modal masks than those for unoccluded objects, which confirms that the proposed method generally performs amodal completion only for occluded objects and not for unoccluded objects. The distribution for occluded objects is also flatter than that for unoccluded objects because the amount by which an object is occluded by can vary significantly from object to object. 

This difference in the two distributions suggests that area ratio can be used to predict the presence or absence of occlusion. We can consider a simple classifier that declares an object to be unoccluded if its area ratio is greater than a threshold. For each value of the threshold, we can compute the precision and recall of this classifier. We plot the precision and recall we obtain by varying this threshold in Fig. \ref{fig:area_ratio_pr_curve}. We compute average precision and find it to be 77.17\%.

\subsection{Direct Evaluation}

To evaluate the accuracy of the mask produced by the proposed method, we need ground truth amodal segmentation annotations. Because no such annotations are publicly available, we collected a set of amodal segmentation masks on 100 objects. For each category, we selected five object instances randomly from the PASCAL VOC 2012 \textit{val} set that are labelled as occluded and annotated them with amodal segmentation masks. For this purpose, we used the annotation tool for MS COCO~\cite{coco}, which is based on the Open Surfaces annotation tool~\cite{opensurfaces}.
\begin{figure}
    \centering
    \subfloat[]{
        \includegraphics[width=0.43\textwidth]{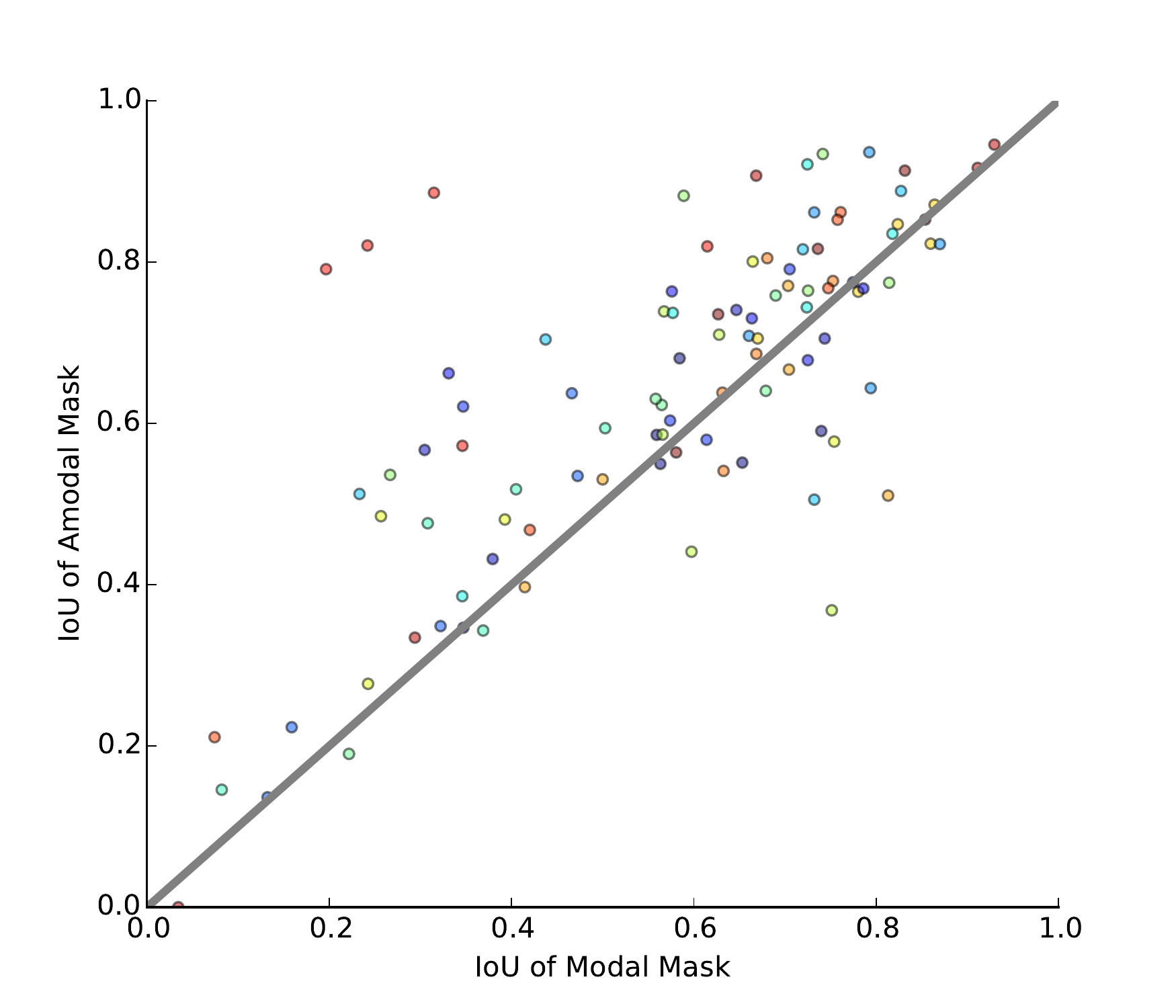}
        \label{fig:overlap_comparison_plot}
    }
    \subfloat[]{
        \includegraphics[width=0.43\textwidth]{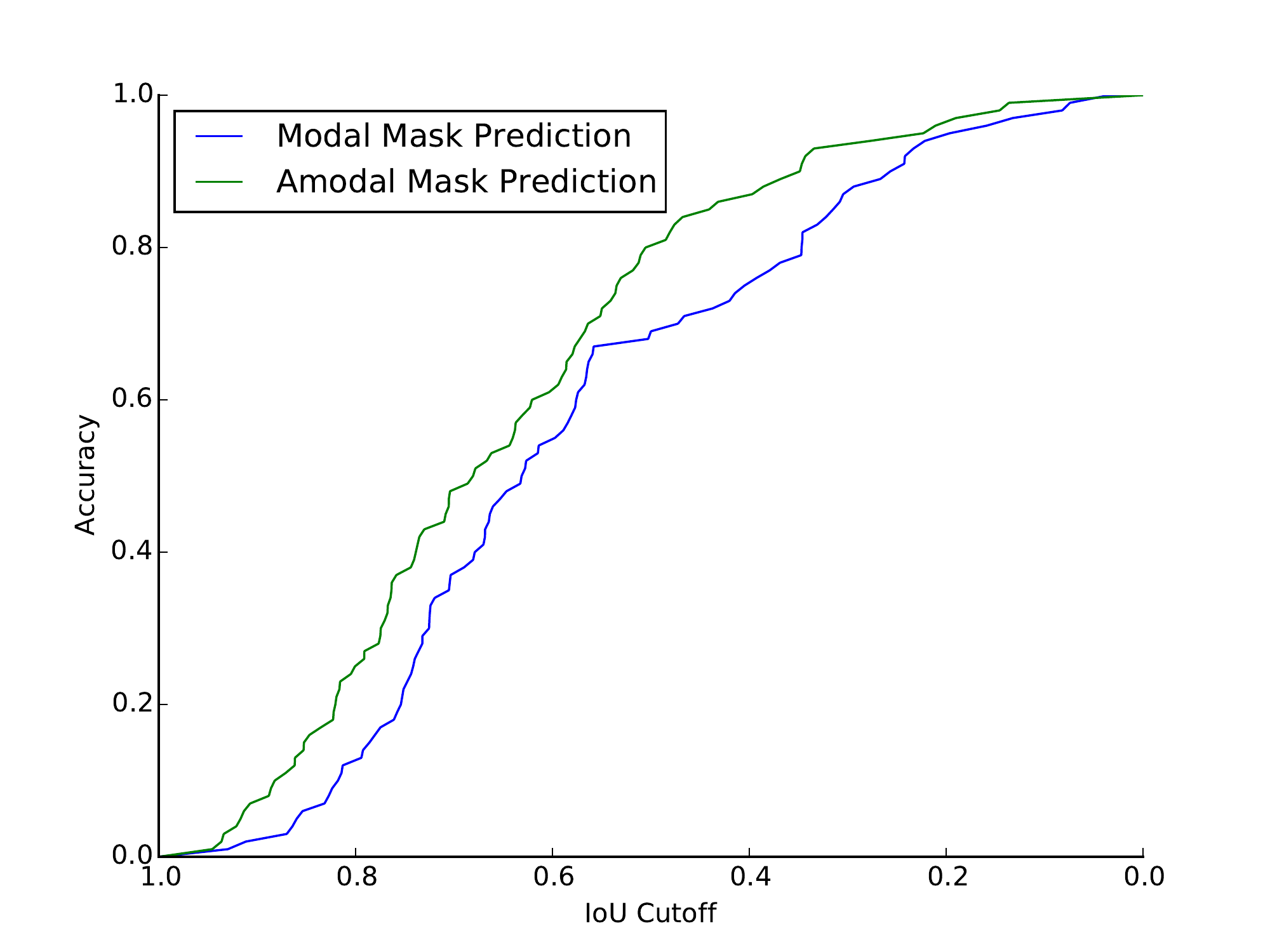}
        \label{fig:overlap_distribution_plot}
    }
    \caption{(a) Comparison of overlap of the modal and amodal mask predictions with the ground truth. Overlap is measured using intersection-over-union (IoU). Each point represents an object instance and the points belonging to object instances in the same category share the same colour. Points that lie above the diagonal represent object instances whose amodal mask predictions are more accurate than their modal mask predictions. (b) Consider the setting where a predicted mask is deemed correct when it overlaps with the ground truth by at least a particular level of cutoff. This plot shows the proportion of object instances whose predicted masks are correct as a function of the cutoff}
\end{figure}

\begin{table}
\centering
\caption{Performance comparison of IIS and the proposed method on the task of amodal instance segmentation}
\footnotesize
\begin{tabular}{l c c c}
\toprule 
Method & Accuracy at 50\% & Accuracy at 70\% & Area Under Curve\\
\midrule
IIS \cite{iis} & $68.0$ & $37.0$ & $57.5$ \\
Proposed Method & $\mathbf{80.0}$ & $\mathbf{48.0}$ & $\mathbf{64.3}$ \\
\bottomrule
\end{tabular}
\label{tab:accu_comp}
\end{table}

We compare the overlap with ground truth achieved by the amodal mask predicted by the proposed method to that achieved by the modal mask predicted by the state-of-the-art modal segmentation method, IIS. In this setting, a modal instance segmentation system represents a fairly strong baseline since in cases where occlusion is not severe, it is possible to omit the occluded portion completely from the predicted mask without significantly lowering intersection-over-union (IoU) with the ground truth. 

\subsubsection{Segmentation Performance}

We first evaluate the segmentation system in isolation by taking the modal bounding box and the category of the object of interest from ground truth. 

As shown in Fig. \ref{fig:overlap_comparison_plot}, on most instances, the masks produced by the proposed method are significantly more accurate than the masks produced by IIS. Notably, the proposed method improves overlap compared to IIS by as much as 20 -- 50\% in many cases. Overall, the proposed method produces better masks than IIS on 73\% of objects. Of the remaining 27\% of objects on which the proposed method performs worse than IIS, the drop in overlap is less than 5\% for the majority of objects. Hence, the masks produced by the proposed method are generally more accurate, sometimes by a sizeable margin. 

In Fig. \ref{fig:overlap_distribution_plot}, we plot the prediction accuracy if all masks with IoUs that exceed a particular cutoff are considered correct. As shown, predicting the mask using the proposed method consistently results in higher accuracy than using IIS at all levels of the cutoff. In Table \ref{tab:accu_comp}, we report the accuracy of the proposed method and IIS at IoU cutoffs of 50\% and 70\%. Additionally, we compute the area under the accuracy curve for both methods. We find that the proposed method performs better than IIS on all metrics. 

\subsubsection{Combined Detection and Segmentation Performance}

Next, we evaluate the performance of the combined detection and segmentation pipeline. We use faster R-CNN~\cite{faster_rcnn} as our detection system and compare overall performance with the proposed method as the segmentation system to IIS. 

We use the amodal segmentation annotations we collected as ground truth and measure performance using mean region average precision ($\mathrm{mAP}^{r}$)~\cite{sds}, which is a common metric used for modal instance segmentation. Region average precision is defined analogously to the standard average precision metric used for detection, except that overlap is computed by finding the pixel-wise IoU between the predicted and the ground truth masks. Because some instances are not annotated with ground truth amodal masks, we are unable to compute region overlap with some ground truth instances. Hence, we make a slight modification to the metric: we use bounding box overlap instead of region overlap to determine which ground truth instance a mask prediction is assigned to. However, we still use region overlap to decide if a mask prediction is deemed correct. 

As shown in Table \ref{tab:pipeline_accu_comp}, the pipeline with the proposed method outperforms the pipeline with IIS by $11.1$ points at $50\%$ overlap and $8.6$ points at $70\%$ overlap. We also include an ablation analysis and report performance on PASCAL 3D+~\cite{xiang2014beyond} annotations in the supplementary material. 

\begin{table}
\centering
\caption{Performance comparison of the combined detection and segmentation pipeline with faster R-CNN as the detection system and either IIS or the proposed method as the segmentation system}
\footnotesize
\begin{tabular}{l c c c}
\toprule 
Method & $\mathrm{mAP}^{r}$ at 50\% IoU & $\mathrm{mAP}^{r}$ at 70\% IoU\\
\midrule
Faster R-CNN \cite{faster_rcnn} + IIS \cite{iis} & $34.1$ & $14.0$ \\
Faster R-CNN \cite{faster_rcnn} + Proposed Method & $\mathbf{45.2}$ & $\mathbf{22.6}$ \\
\bottomrule
\end{tabular}
\label{tab:pipeline_accu_comp}
\end{table}

\section{Conclusion}

We presented a new method for amodal instance segmentation, which represents the first such method to the best of our knowledge. We introduced a novel strategy for generating synthetic amodal instance segmentation data from modal instance segmentation annotations. This strategy enabled us to train a model for amodal instance segmentation despite the the lack of publicly available amodal segmentation data. Additionally, we presented a new approach for iteratively predicting the amodal bounding box from amodal segmentations. We demonstrated the effectiveness of the proposed method in predicting amodal segmentation masks both qualitatively and quantitatively. 
\\

\scriptsize{
\noindent\textbf{Acknowledgements.}
This work was supported by ONR MURI N00014-09-1-1051 and ONR MURI N00014-14-1-0671. Ke Li thanks the Natural Sciences and Engineering Research Council of Canada (NSERC) for fellowship support. The authors also thank Saurabh Gupta and Shubham Tulsiani for helpful suggestions and NVIDIA Corporation for the donation of GPUs used for this research. }
\clearpage

\normalsize
\bibliographystyle{splncs03}
\bibliography{amodal}

\clearpage

\title{Amodal Instance Segmentation} 
\subtitle{Supplementary Material}

\titlerunning{Amodal Instance Segmentation}

\authorrunning{Ke Li and Jitendra Malik}

\author{Ke Li and Jitendra Malik}


\institute{Department of Electrical Engineering and Computer Sciences,\\
	University of California, Berkeley\\
	\email{ \{ke.li,malik\}@eecs.berkeley.edu}
}

\maketitle

\section{Ablation Analysis}

We analyze the incremental effects of changing various components of the training procedure. To this end, we trained several variants of the model: in one variant, the model takes an amodal heatmap rather than a modal heatmap as input, where the amodal heatmap is produced by a model with the same architecture as IIS, but is trained to predict the amodal segmentation masks directly. In a different variant, we only generate a fixed occlusion configuration per object rather than many different occlusion configurations that vary identities, positions and scales of occluding objects. We will refer to the first variant as the ``model without modal segmentation prediction'' and the second variant as the ``model without dynamic sample generation''. We report the performance of the pipelines with these variants on the dataset we collected and compare to the performance of the original model in Table \ref{tab:pipeline_accu_comp_ablation}. As shown, the original model largely performs better than these variants, suggesting that feeding in modal segmentation prediction as input and generating diverse occlusion configurations dynamically are both important ingredients to achieving good performance. 

\begin{table}
\centering
\caption{Performance comparison of the combined detection and segmentation pipeline with faster R-CNN as the detection system and different variants of the proposed method as the segmentation system}
\footnotesize
\begin{tabular}{l c c c}
\toprule 
Method & $\mathrm{mAP}^{r}$ at 50\% IoU & $\mathrm{mAP}^{r}$ at 70\% IoU\\
\midrule
Without Modal Segmentation Prediction & $35.2$ & $18.4$ \\
Without Dynamic Sample Generation & $39.8$ & $\mathbf{22.7}$ \\
With Both & $\mathbf{45.2}$ & $22.6$ \\
\bottomrule
\end{tabular}
\label{tab:pipeline_accu_comp_ablation}
\end{table}

\section{Results on PASCAL 3D+}

We also evaluate performance on the PASCAL 3D+ dataset~\cite{xiang2014beyond}, which provides CAD models aligned to rigid objects in PASCAL images. The projections of CAD models onto the image plane can be treated as amodal segmentation masks~\cite{kar2015amodal}. Due to the mismatch between the shape of CAD models and that of objects depicted in images, these masks are only approximately correct, however. We report region average precision on the subset of rigid objects in the PASCAL VOC 2012 \textit{val} set for which CAD models are available in Table \ref{tab:pipeline_accu_comp_pascal3d}. 

\begin{table}
\centering
\caption{Performance comparison of the combined detection and segmentation pipeline on rigid objects and their corresponding annotations from PASCAL 3D+}
\footnotesize
\begin{tabular}{l c c c}
\toprule 
Method & $\mathrm{mAP}^{r}$ at 50\% IoU & $\mathrm{mAP}^{r}$ at 70\% IoU\\
\midrule
Faster R-CNN \cite{faster_rcnn} + IIS \cite{iis} & $37.4$ & $15.9$ \\
Faster R-CNN \cite{faster_rcnn} + Proposed Method & $\mathbf{44.0}$ & $\mathbf{20.9}$ \\
\bottomrule
\end{tabular}
\label{tab:pipeline_accu_comp_pascal3d}
\end{table}

\section{Additional Visualizations}

We provide additional examples of modal and amodal heatmap and segmentation mask predictions produced by the proposed method and IIS~\cite{iis}. 

\begin{figure}
    \centering
    \includegraphics[width=1.0\textwidth]{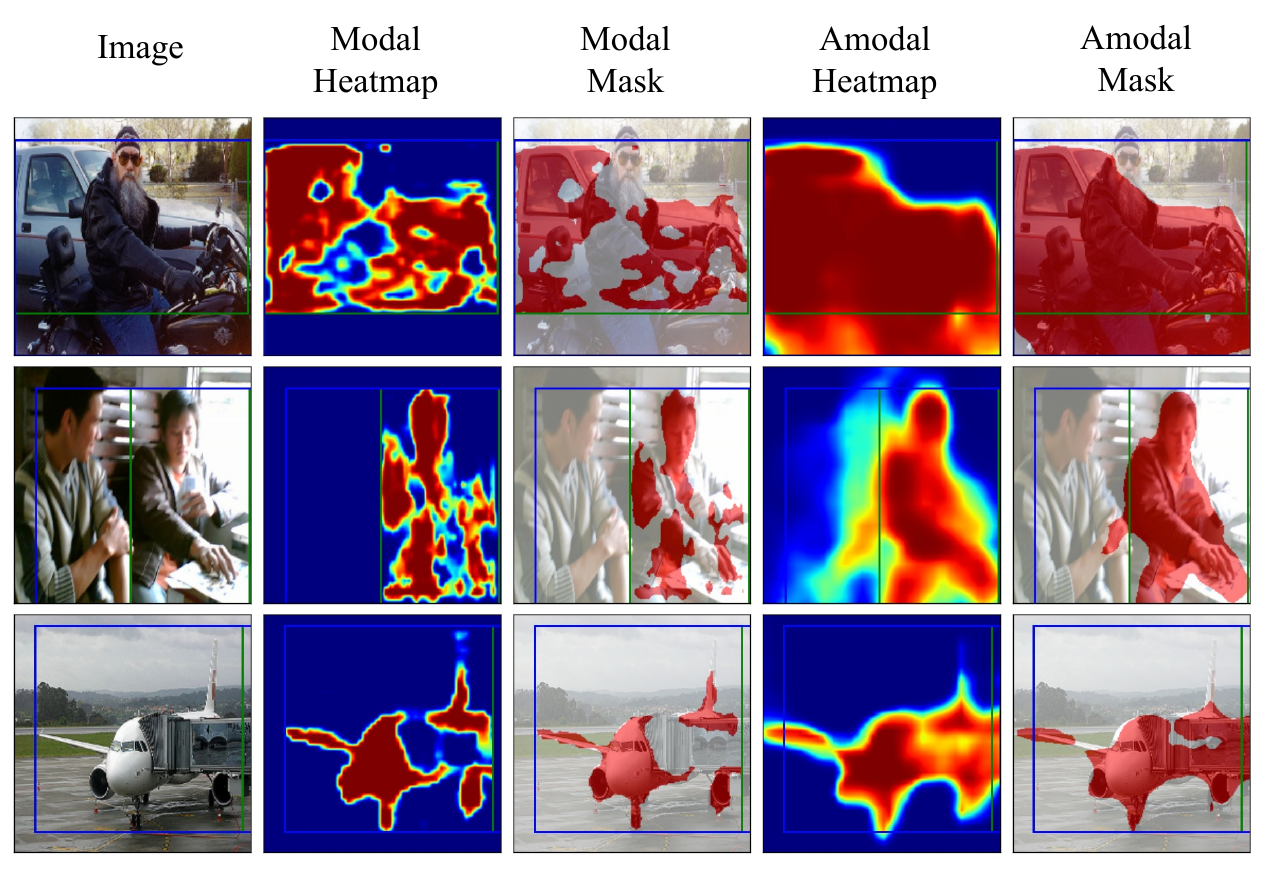}
    \caption{Additional examples of modal and amodal segmentation heatmap and segmentation mask predictions}
    \label{fig:supp5}
\end{figure}

\begin{figure}
    \centering
    \includegraphics[width=1.0\textwidth]{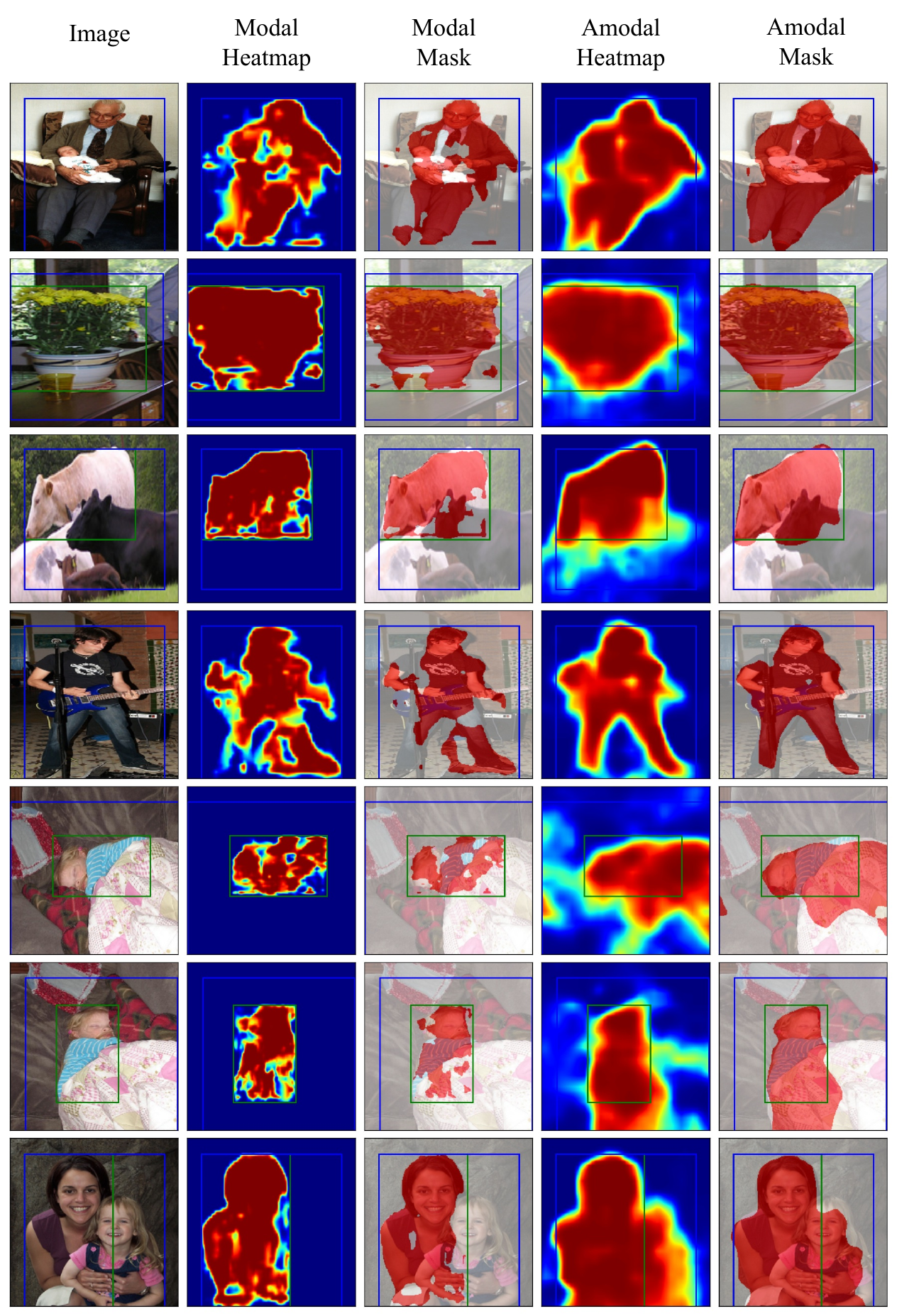}
    \caption{Additional examples of modal and amodal segmentation heatmap and segmentation mask predictions}
    \label{fig:supp4}
\end{figure}

\begin{figure}
    \centering
    \includegraphics[width=1.0\textwidth]{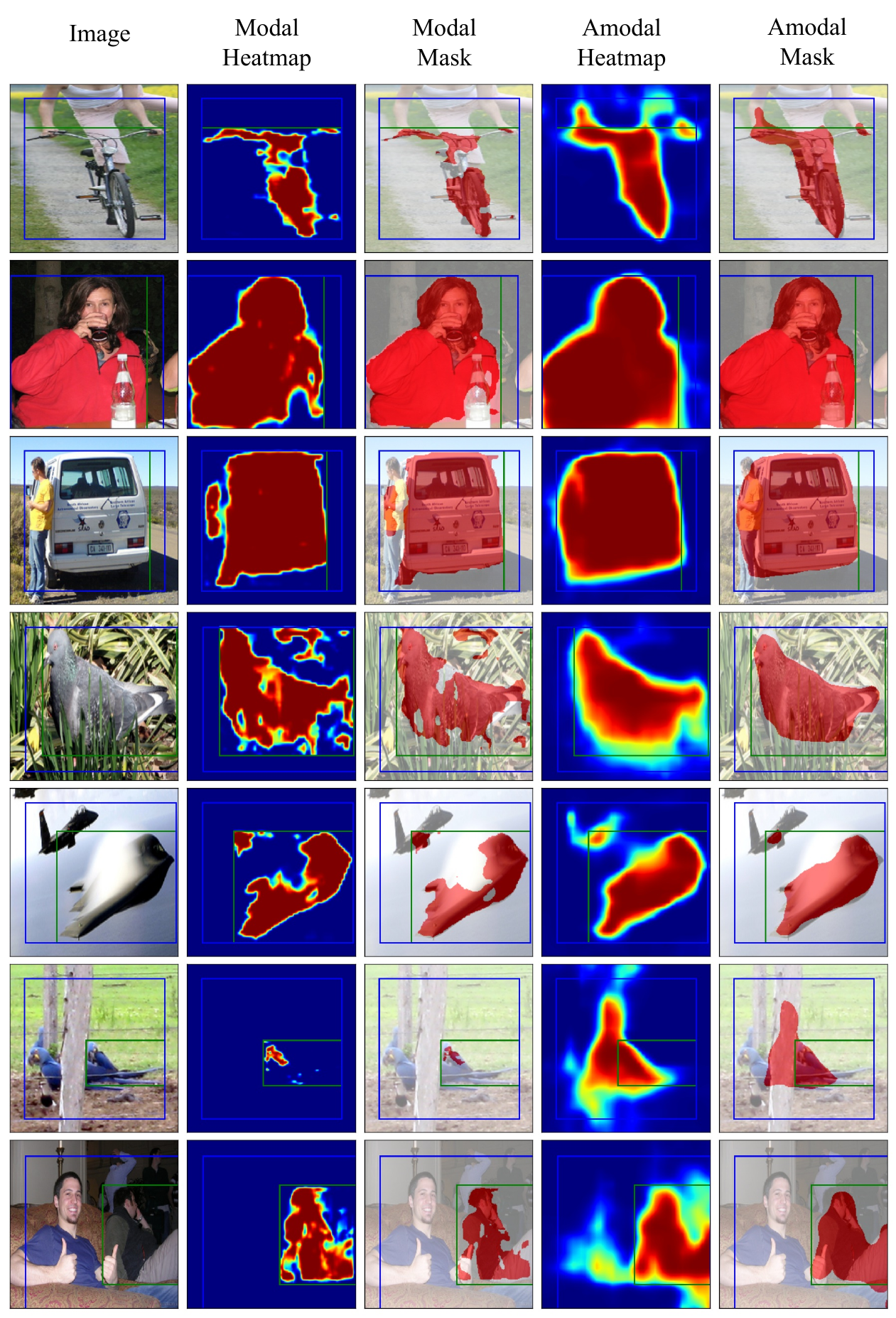}
    \caption{Additional examples of modal and amodal segmentation heatmap and segmentation mask predictions}
    \label{fig:supp1}
\end{figure}

\begin{figure}
    \centering
    \includegraphics[width=1.0\textwidth]{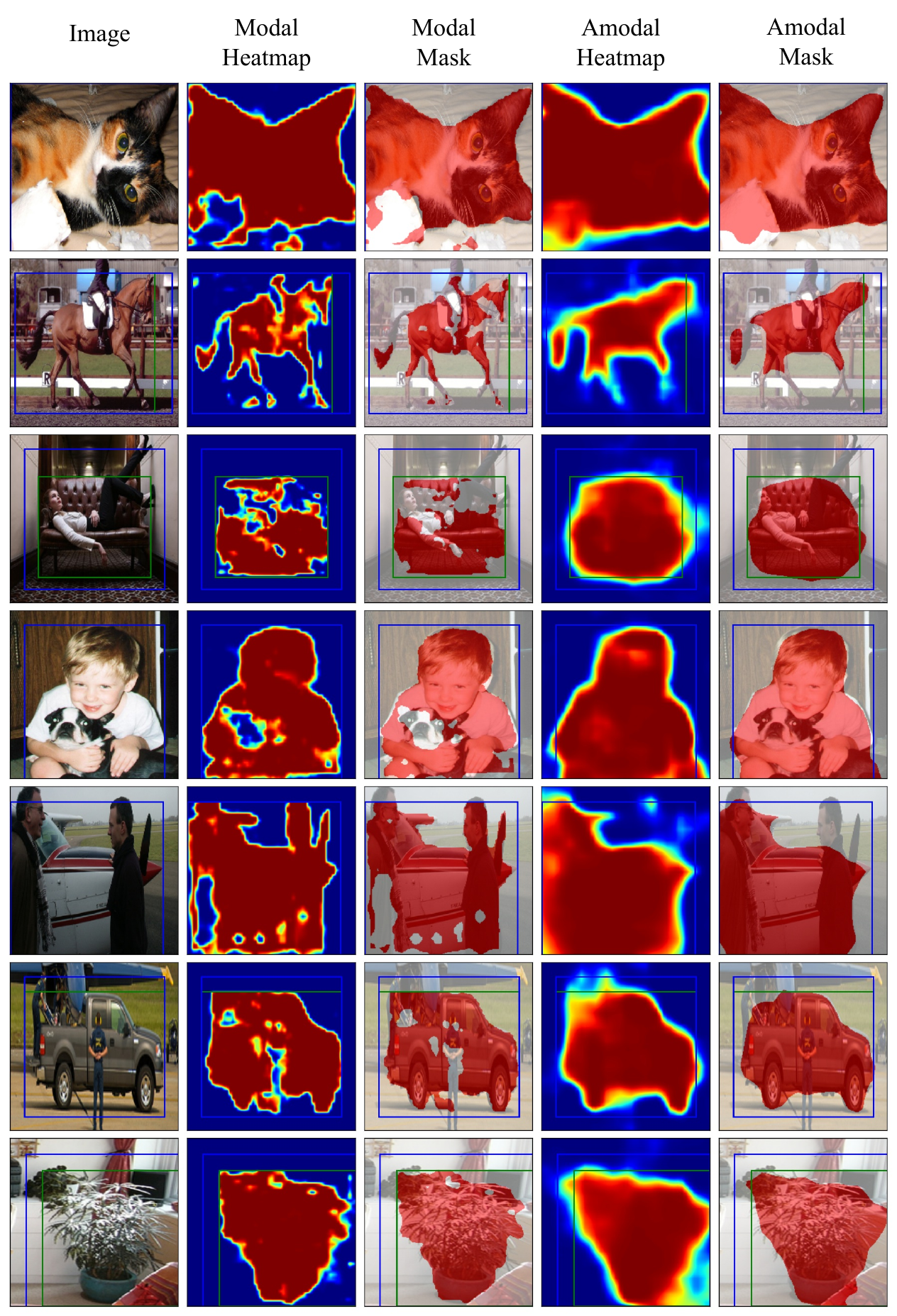}
    \caption{Additional examples of modal and amodal segmentation heatmap and segmentation mask predictions}
    \label{fig:supp2}
\end{figure}

\begin{figure}
    \centering
    \includegraphics[width=1.0\textwidth]{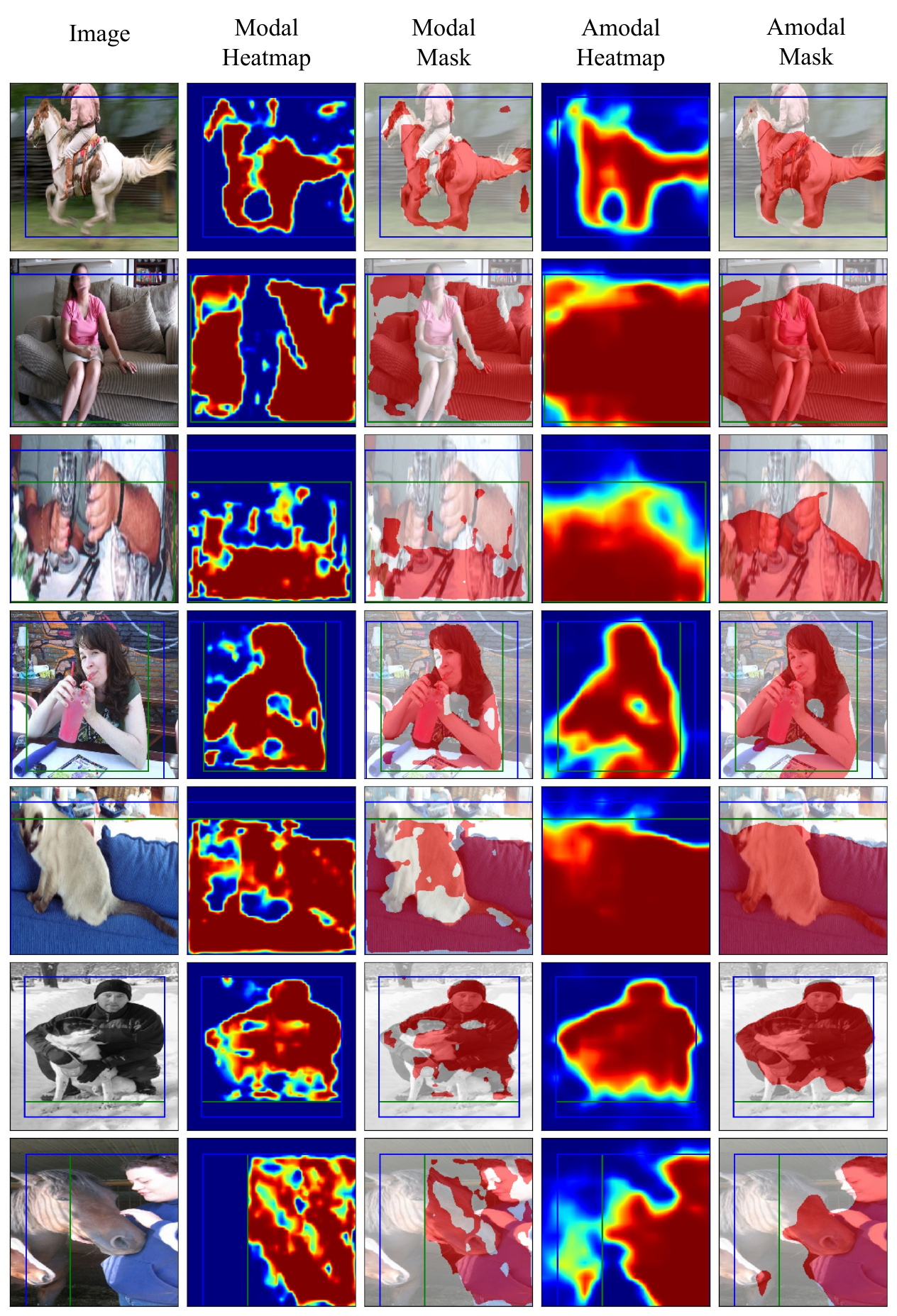}
    \caption{Additional examples of modal and amodal segmentation heatmap and segmentation mask predictions}
    \label{fig:supp3}
\end{figure}

\end{document}